# Probabilistic spatial clustering based on the Self Discipline Learning (SDL) model of autonomous learning

Zecang Gu[a], Xiaoqi Sun[a], Yuan Sun[a], Fuquan Zhang[b,*]
[a]Apollo Japan Co., Ltd.
[b]College of Computer and Control Engineering, Minjiang University

**Abstract**: Unsupervised clustering algorithm can effectively reduce the dimension of high-dimensional unlabeled data, thus reducing the time and space complexity of data processing. However, the traditional clustering algorithm needs to set the upper bound of the number of categories in advance, and the deep learning clustering algorithm will fall into the problem of local optimum. In order to solve these problems, a probabilistic spatial clustering algorithm based on the Self Discipline Learning(SDL) model is proposed. The algorithm is based on the Gaussian probability distribution of the probability space distance between vectors, and uses the probability scale and maximum probability value of the probability space distance as the distance measurement judgment, and then determines the category of each sample according to the distribution characteristics of the data set itself. The algorithm is tested in Laboratory for Intelligent and Safe Automobiles(LISA) traffic light data set, the accuracy rate is 99.03%, the recall rate is 91%, and the effect is achieved.

## 1 Introduction

Clustering is a process of grouping data according to the similarity of samples, which is a basic unsupervised learning task with many applications. The measurement of similarity or difference between samples plays an important role in data clustering. Specifically, similarity or difference is determined by data description and scale. On the basis of similarity measurement, various clustering rules are developed. These methods include spatial partition based methods (such as k-means [1-3], spectral clustering [4]) and hierarchical methods (such as birch [5]). Among them, K-means algorithm is widely used because of its high processing efficiency for large data sets. There are many advantages and disadvantages of this algorithm: firstly, it only considers the intra class distance, but does not consider the inter class distance; secondly, the determination of the upper bound of the number of clusters of the data sets containing a large number of samples mainly depends on experience, while the upper bound of the number of clusters set artificially is often too large, which leads to the reduction of the efficiency of the algorithm.

When the data dimension increases to a large extent, the ability of distance measurement will be weakened, so researchers have developed more stable measurement methods, such as metric learning method [29] and dimension reduction method [8, 17, 18]. More specifically, [9] studied the relationship between the K-means clustering target and the principal components extracted by PCA, and the experiments verified the effectiveness of using PCA for dimensionality reduction and K-means clustering. Similarly, in [10], LDA is combined with PCA and K-means in low dimensional subspace to further reduce the dimension and select the most suitable clustering subspace. Finally, a provable and accurate feature selection method is proposed in [7], which is based on singular value decomposition and random projection. Because the original data can not be determined after dimension reduction, it is not convenient for subsequent analysis.

With the development of deep learning technology, researchers have been committed to using deep neural network for joint representation learning and clustering, namely deep clustering. As an unsupervised tool, automatic encoder is used for dimensionality reduction of nonlinear feature extraction, classification and clustering tasks. It learns general features by optimizing the cleverly designed objective functions of some pre-tasks. All supervised pseudo tags are generated automatically from input images without manual annotation. Common pre tasks include image completion [11], image coloring [12], jigsaw [13], counting [14], rotation [15], clustering [16]. Although great progress has been made, deep clustering still performs poorly in natural images (such as ImageNet) compared with simple handwriting. In order to improve the accuracy of image recognition, the feature vectors of images are clustered according to the actual results of Gaussian distribution, which is the inherent probability space of data. In this paper, a probabilistic space clustering method based on self-learning is proposed, which can avoid the weakening phenomenon of dimension increase by calculating the distance of probability space. Combined with the Self Discipline Learning (SDL) model[21-22], the algorithm does not need to set the upper bound of the number of categories in advance, and reduces the risk of falling into the local optimal solution. It is applied to traffic light recognition and achieves good results.

## 2 Methodology theory

Before introducing the probabilistic spatial clustering algorithm based on the SDL model, we need to introduce several definitions in order to read this paper better. Probability space distance is the Euclidean distance of two probability space centers minus two probability space scales. The distance from the feature vector to the probability space is the Euclidean distance from the feature vector to the center of the probability space minus the probability scale of the probability space.

### 2.1 Probability Space

The probability space described here is based on the Soviet mathematician Andrey Kolmogorov's theory that "probability theory is based on measurement theory". The so-called probability space is a measurable space with a total measure of "1". According to this theory, lemma 1 can be generated: "there is only one probability distribution in probability space, so there are infinite probability spaces in Euclidean space. According to the above theorem of probability space, there is lemma 2: "the space distance with probability distribution of "1" in probability space is "0". According to lemma 2, we can define a distance formula across Euclidean space and probability space.

As shown in Fig. 1, in Euclidean space E, there exists a probability space $\mathcal{G}^{(w_j)}$ whose center of probability distribution is $\mathcal{W}_j \in W(j = 1,2,\ldots,n)$, and there is a point $\gamma_j \in \mathcal{R}(j = 1,2,\ldots,n)$ between it and the center of probability distribution $\mathcal{V}_j \in V(j = 1,2,\ldots,n)$ in another probability space. Here, W, V and R are the set of Euclidean space or probability space respectively. Which one is closer to the set W, V or R? This is also the most typical problem of pattern recognition. It is not difficult to see that as long as the distance between W set and V set is solved, such a proposition is not difficult to solve.

The classification of artificial intelligence must be based on the distance between Euclidean space and probability space as shown in Figure 1.

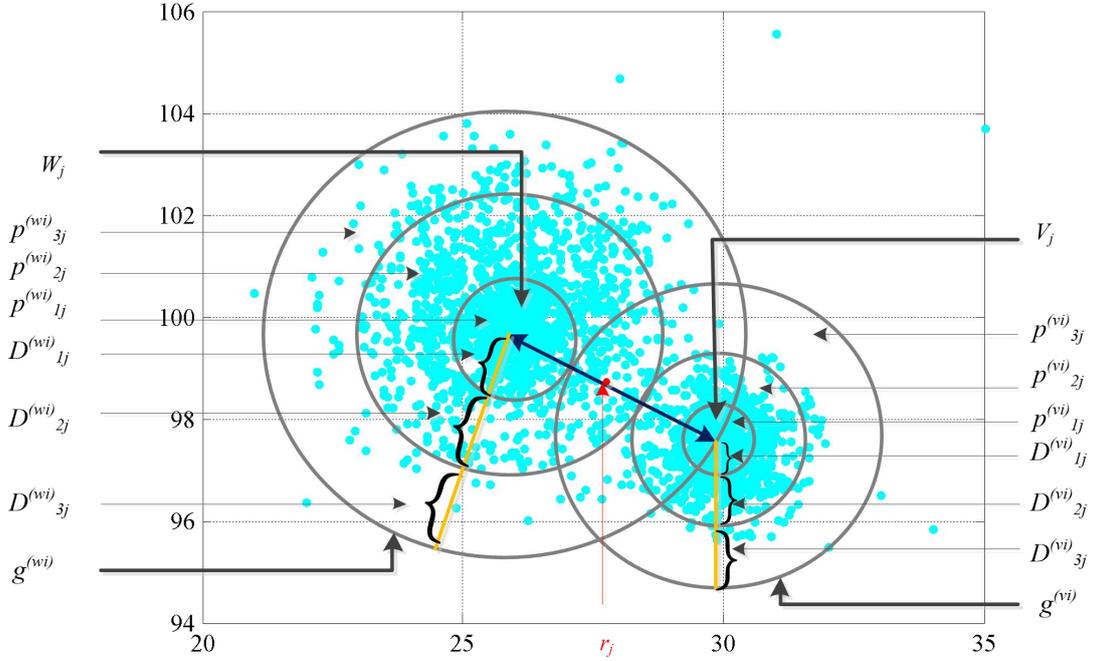

Fig. 1 Schematic diagram of optimal classification in probability space

Let $\mathcal{D}_{ij}^{(w_j)}$ be the scale of the probability distribution center $\mathcal{W}_j \in W$, the Euclidean space distance between $(i)$ and $(i+1)$, and the probability distribution value $\rho_{ij}^{(w_j)}$ ($i = 1, 2, \ldots m_j^{(w_j)}$) in the region of this scale. Then the distance between the set V of probability space $\mathcal{G}^{(v_j)}$ and the set W of probability space $\mathcal{G}^{(w_j)}$ can be calculated by the following formula:

$$G(V,W) = \sqrt{\sum_{j=1}^{n}(r_j - v_j)^2} + \sqrt{\sum_{j=1}^{n}(r_j - w_j)^2} = \sqrt{\sum_{j=1}^{n}(v_j - w_j)^2} = \sqrt{\sum_{j=1}^{n}(w_j - v_j)^2}$$

（1）

$$(r_j - v_j) = \begin{cases} 0 & |r_j - v_j| \leq \Delta_j^{(V_j)} \\ |r_j - v_j| - \Delta_j^{(V_j)} & |r_j - v_j| > \Delta_j^{(V_j)} \end{cases}$$

$$(r_j - w_j) = \begin{cases} 0 & |r_j - w_j| \leq \Delta_j^{(W_j)} \\ |r_j - w_j| - \Delta_j^{(W_j)} & |r_j - w_j| > \Delta_j^{(W_j)} \end{cases}$$

$$(r_j - w_j) = (w_j - v_j) = \begin{cases} 0 & |r_j - w_j| \le (\Delta_j^{(V_j)} + \Delta_j^{(W_j)}) \\ |r_j - w_j| - (\Delta_j^{(V_j)} + \Delta_j^{(W_j)}) & |r_j - w_j| > (\Delta_j^{(V_j)} + \Delta_j^{(W_j)}) \end{cases} \quad (2)$$

$$\Delta_j^{(V_j)} = \sum_{i=1}^{m_j^{(V_j)}} D_{ij}^{(V_j)} \rho_{ij}^{(V_j)}$$

$$\Delta_j^{(W_j)} = \sum_{i=1}^{m_j^{(W_j)}} D_{ij}^{(W_j)} \rho_{ij}^{(W_j)} \quad (3)$$

In the above formula, $(\Delta_j^{(V_j)} + \Delta_j^{(W_j)})$ is in the probability space $\mathcal{G}^{(V_j)}$ and $\mathcal{G}^{(W_j)}$. Since the distance between the probability space in the area with probability distribution of "1" should be "0", it is the error value between the Euclidean distance and the distance of the probability space. By eliminating these two error values, we can get the strict uniform distance between the probability space $\mathcal{G}^{(V_j)}$ and $\mathcal{G}^{(W_j)}$, which can unify the distance between Euclidean space and probability space. The above formulas (1) - (3) can satisfy the distance scale conditions of nonnegativity, non degeneracy, symmetry and trigonometric inequality. Figure 2-(a) explains how the probability space distances obtained by clustering satisfy the trigonometric inequality.

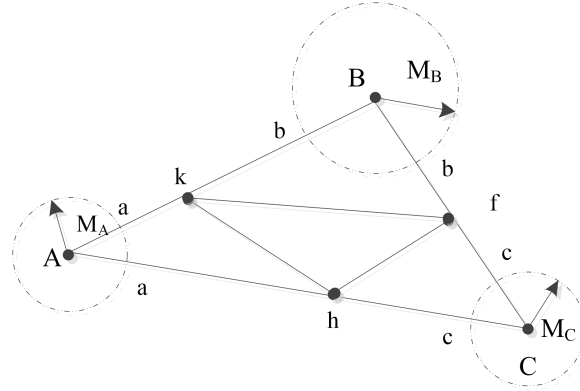

(a)

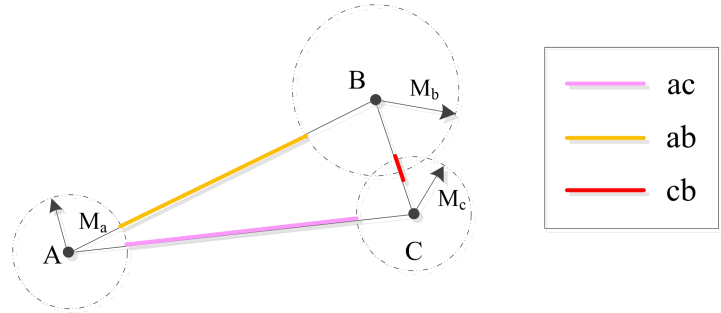

(b)

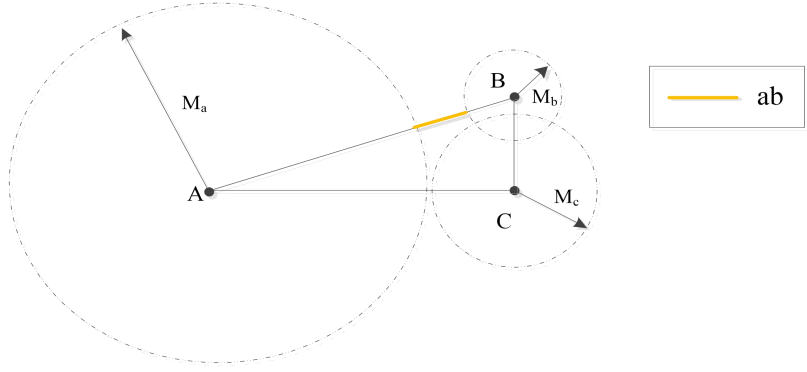

(c)

Fig. 2 Schematic diagram of trigonometric inequality

Firstly, three probability spaces a, B and C are selected. The maximum probability values of the three probability spaces are A(x1, Y1), B(X2, Y2), and C(X3, Y3). The maximum probability scales are $M^A$, $M^B$, and $M^C$, respectively.

$$x_k = \frac{1}{2}[(x_1 + M_A) + (x_2 - M_B)] \quad x_1 < x_2 \quad (4)$$

If $|x_1 - x_2| \leq M_A$ or $|x_1 - x_2| \leq M_B$, then $M_A = 0, M_B = 0$.

$$y_k = \frac{1}{2}[(y_1 + M_A) + (y_2 - M_B)] \quad y_1 < y_2 \quad (5)$$

If $|y_1 - y_2| \leq M_A$ or $|y_1 - y_2| \leq M_B$, then $M_A = 0, M_B = 0$.

$$x_f = \frac{1}{2}[(x_2 + M_B) + (x_3 - M_C)] \quad x_2 < x_3 \quad (6)$$

If $|x_2 - x_3| \leq M_B$ or $|x_2 - x_3| \leq M_C$, then $M_B = 0, M_C = 0$.

$$y_f = \frac{1}{2}[(y_2 - M_B) + (y_3 + M_C)] \quad y_2 \geq y_3 \quad (7)$$

If $|y_2 - y_3| \leq M_B$ or $|y_2 - y_3| \leq M_C$, then $M_B = 0, M_C = 0$.

$$x_h = \frac{1}{2}[(x_3 - M_C) + (x_1 + M_A)] \quad x_3 \geq x_1 \quad (8)$$

If $|x_3 - x_1| \leq M_C$ or $|x_3 - x_1| \leq M_A$, then $M_C = 0, M_A = 0$.

$$y_h = \frac{1}{2}[(y_3 + M_B) + (y_1 - M_C)] \quad y_3 < y_1 \quad (9)$$

If $|y_3 - y_1| \leq M_C$ or $|y_3 - y_1| \leq M_A$, then $M_C = 0, M_A = 0$.

(Kx, Ky), (Fx, Fy), (Hx, Hy) are formed into a triangle, which must satisfy the trigonometric inequality. On this basis, two additional cases are added:

In Figure 2-(b), there is an edge "0": when ab ≠ ac and cb = 0, then AB = AC = ab ∩ ac.

In Figure 2-(c), there is only one side: AB = AC = ab.

## 2.2 Clustering based on SDL model

As shown in Figure 3, the clustering method is summarized and applied to the LISA traffic light dataset [19-20] to improve the accuracy of traffic light recognition.

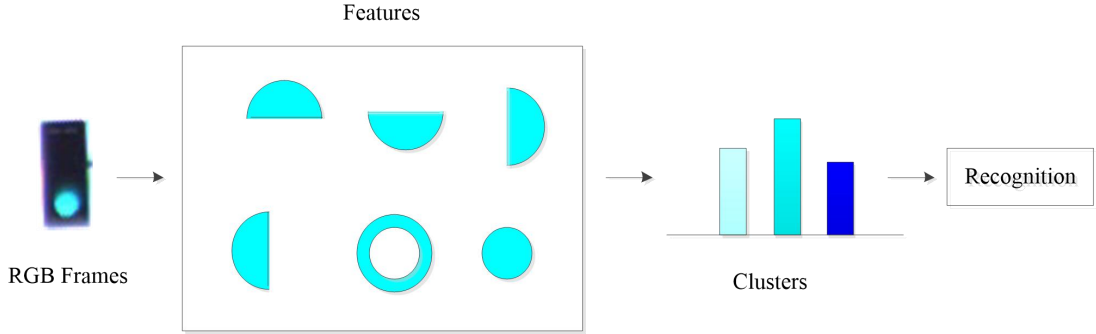

Fig. 3 Overview of traffic light recognition based on the SDL clustering

The initial step of step 1: take part of the image in LISA traffic light dataset and extract 18 feature values of the traffic light image as a feature vector. The maximum probability value of each feature vector is calculated, and the whole feature vector data is divided into K initial regions according to the modulus of the maximum probability value of each feature vector.

Find the maximum probability space of step2: set the maximum number of migration (3 times) and the maximum number of convergence (1 time), and use the Self Discipline Learning (SDL) model [21-22] and the distance of probability space (formula 1-3) to calculate K regions respectively. For example, the maximum probability space of the respective feature vector data of K regions at the initial time.

Correct the error of cluster calculation by modulus of step3: take the maximum probability space of the feature vectors of two adjacent regions of K regions as the center, and take the distance scale of probability space as the data selection benchmark, carry out the data exchange between the two regions, so that the feature vectors of each region gather around the maximum probability space and become a class, so as to offset the error of classification due to the size of the module.

Judgment of step4: judge whether the number of clusters is close to the demand. If yes, the maximum probability space of each region in the cluster area is calculated and the clustering is finished. Otherwise, each region is divided into two regions by the size of module to form a new cluster number kn, and the process of step2 is repeated.

Here the maximum probability space is the Gaussian distribution obtained by the SDL model. The specific steps of probabilistic spatial clustering algorithm based on the SDL model are shown in Table 1.

Table 1 **SDL clustering algorithm**

**Input:** V(C) (h=1,2,...,$\rho$)　All feature values in a given region C
**Output:** C(k) （k=$2^1$, $2^2$,... $2^n$, ）
　for i ← 0 to n **do**
　　for m ← 0 to μ **do**
　　　A{$G^{(m)}$[$V(C)$]}←
　　　G{ A{$G^{(m)}$[$V(C)$]},M{$G^{(m)}$[$V(C)$],$A^{(m)}$[$V(C)$]}
　　　**if**[$A^{(m)} - A^{(m+1)}$]$^2 \leq \delta$ **then**
　　　　break

```
            end if
        end for
      for ϵ ← 0 to 2^i do
        for m ← 0 to μ do
            G^(m+1) [C^(ε)] ← G{A[G^(m) [C^(ε)]],M[G^(m) [C^(ε)],A^(m) [C^(ε)]]}
             if[A^(m) – A^(m+1)]^2 ≤ δ then
                 break
             end if
        end for
        for m ← 0 to h do
            if Dp^(m) {V^(m) [C^(ε)],A^(m) ) [C^(ε)]}< Dp^(m) {V^(m) [C^(ε+1)],A^(m) ) [C^(ε+1)]} then
                [C^(ε)] ← V^(m) [C^(ε)]
            end if
            if Dp^(m) {V^(m) [C^(ε)],A^(m) ) [C^(ε)]} ≥ Dp^(m) {V^(m) [C^(ε+1)],A^(m) ) [C^(ε+1)]} then
                [C^(ε*1)] ← V^(m) [C^(ε)]
            end if
        end for
      end for
   end for

 Notes: A---Maximum probability value
        M---Probability scale
        G---Probability space
```

After obtaining all probability spaces, we hope to show the position of each probability space in the graph, so as to judge the correctness of the probability space. Through the analysis of two-dimensional spatial images, we can find that the maximum probability value of some probability spaces is included in other probability spaces, but the actual data show that the probability spaces after clustering do not intersect each other. In fact, the position of each probability space can not be displayed in a two-dimensional space, but projected in two-dimensional coordinates. As shown in Figure 4, the length of the red dotted line is the projection value, so it's concluded that the position of the probability space exists in the multidimensional space.

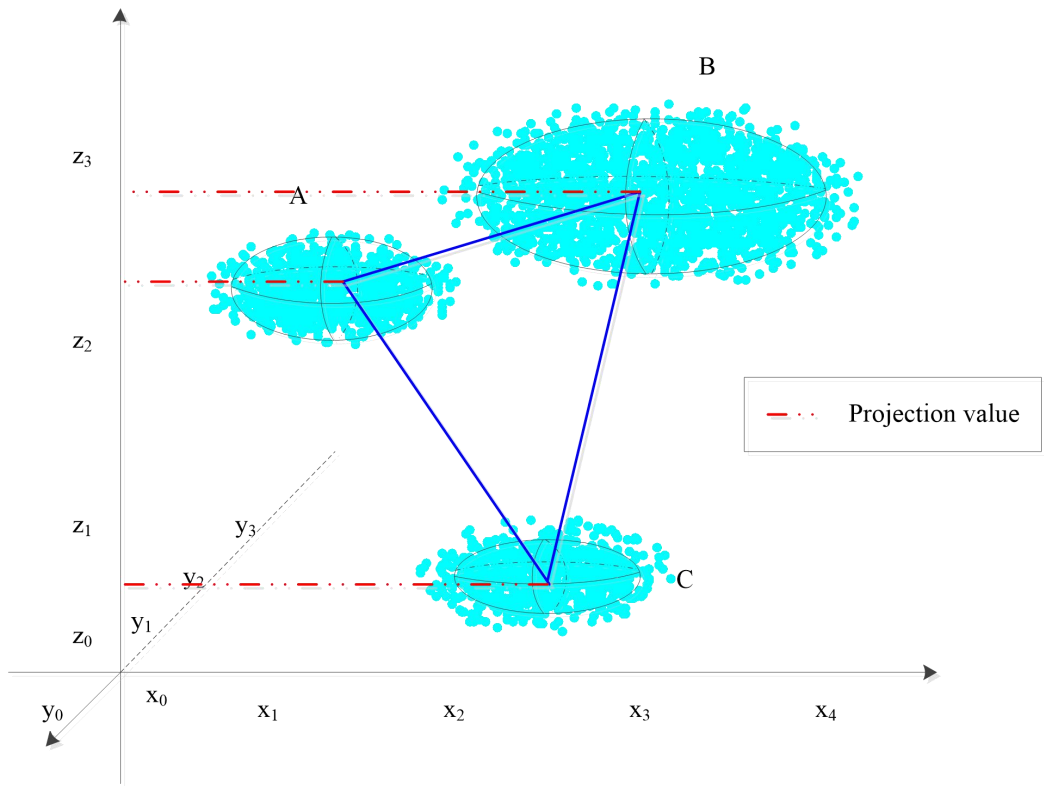

Fig. 4 Spatial projection

Figure 5 is an example of a simple probabilistic spatial clustering algorithm based on the SDL model. It shows that the pixel value features of red and green traffic light areas constitute multiple feature vectors, and behavior samples are listed as feature values. The method of three translation and one convergence is used to cluster. The data processed by step1-step3 are shown in Figure (a) - (b); after step 4, the space with probability space less than zero is merged, as shown in Figure (c).

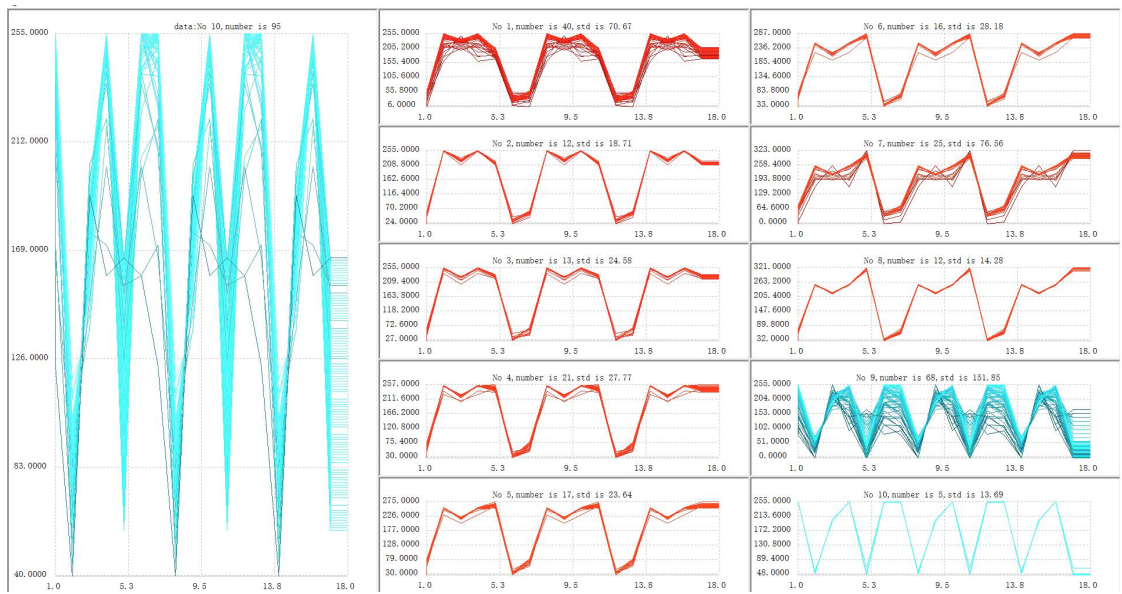

(a)

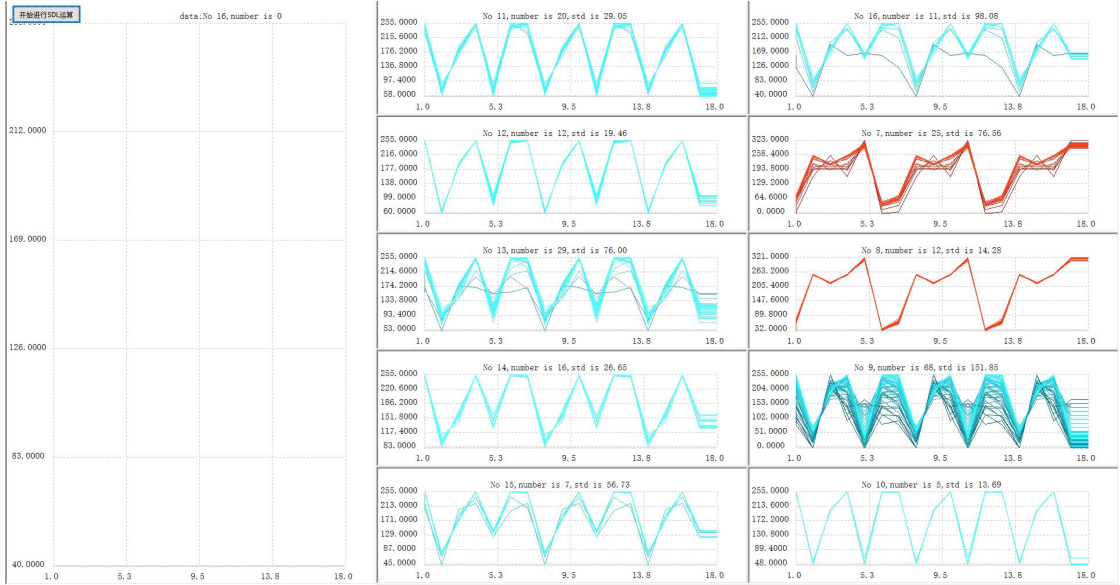

(b)

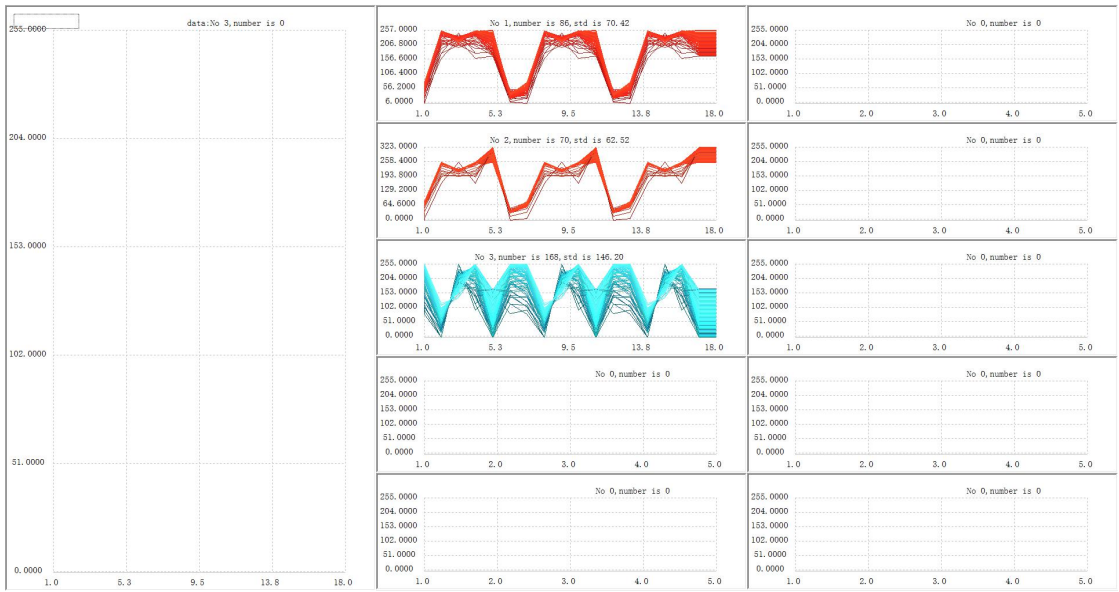

(c)

Fig. 5 Application of probabilistic spatial clustering algorithm based on the SDL model

## 3 Analysis and results

In order to detect and recognize traffic lights in real time and effectively, traditional image processing method and deep learning method of convolution neural network are often used. George[23] adopted the fast radial symmetry transformation method to detect the circular area(its main feature is to detect symmetrical spots in the image, and produce local maximum in the center of bright spot and local minimum in the center of dark spot). The input image is 640 * 480, 8-bit RGB color image. The running time is 3.26 seconds, the accuracy rate is 53.5%, and the recall rate is 59.15%. Sooksatra[24] improved the method in [25], which only detects red and yellow lights, and extracts 55% of the top part of the image as the input region of interest. The input image size is 240 * 320, the running time is 1.82 seconds, the accuracy rate is 84.93%, and the recall rate is 87.32%. Haltakov[26] used semantic segmentation method to generate traffic light candidate areas, and used the color, texture and geometric features of traffic lights to detect traffic lights. The accuracy rate of traffic lights in French weather was 73.2%, and the recall rate was 71.7%, while that of German weather was 71.5% and 84.3%. Zhang[27] obtained the street video of several cities in China, used the top hat method to operate the V-channel of HSV image, and then recognized the traffic lights according to the circular light and black frame, with an accuracy of 85%. Lu[28] used the concept of visual attention mechanism and based on Fast RCNN network, predicted the candidate area where the signal lamp was located, and then determined the final position of the signal lamp, and obtained an average accuracy rate of more than 80.0% on the LISA traffic light dataset. This method greatly reduces the search space on the high-resolution image and improves the detection speed, but the recognition accuracy is not so good. Jensen[29] detected the daytime traffic lights in LISA traffic light dataset based on yolov3, and identified the accurate recall curve area up to 90.49%. Ouyang[30] used heuristic candidate region selection module and convolutional neural network to identify traffic lights. The accuracy rate on LISA traffic light dataset reached 96.6%, and the recall rate was 66.6%. Although these methods improve the accuracy, there are serious problems of missed detection.

  In this paper, the daytime data set of LISA traffic light dataset is used. The video duration is 44 minutes and 41 seconds, and 43007 frames of images are captured at the resolution of 1280 × 960. There are 118331 manually marked traffic lights, and there are 7 kinds of label categories, which are red yellow green three kinds of round signal lights, red yellow green three kinds of left arrow signals, and green forward arrow signals. However, due to the uneven distribution of samples in each category, only go and stop categories are set in reference paper [28], and the original data are annotated again. Due to the small pixel size of the signal lamp in the image, in order to increase the proportion of the signal lamp area in the image, the image is divided into 640 × 480, as shown in Figure 6.

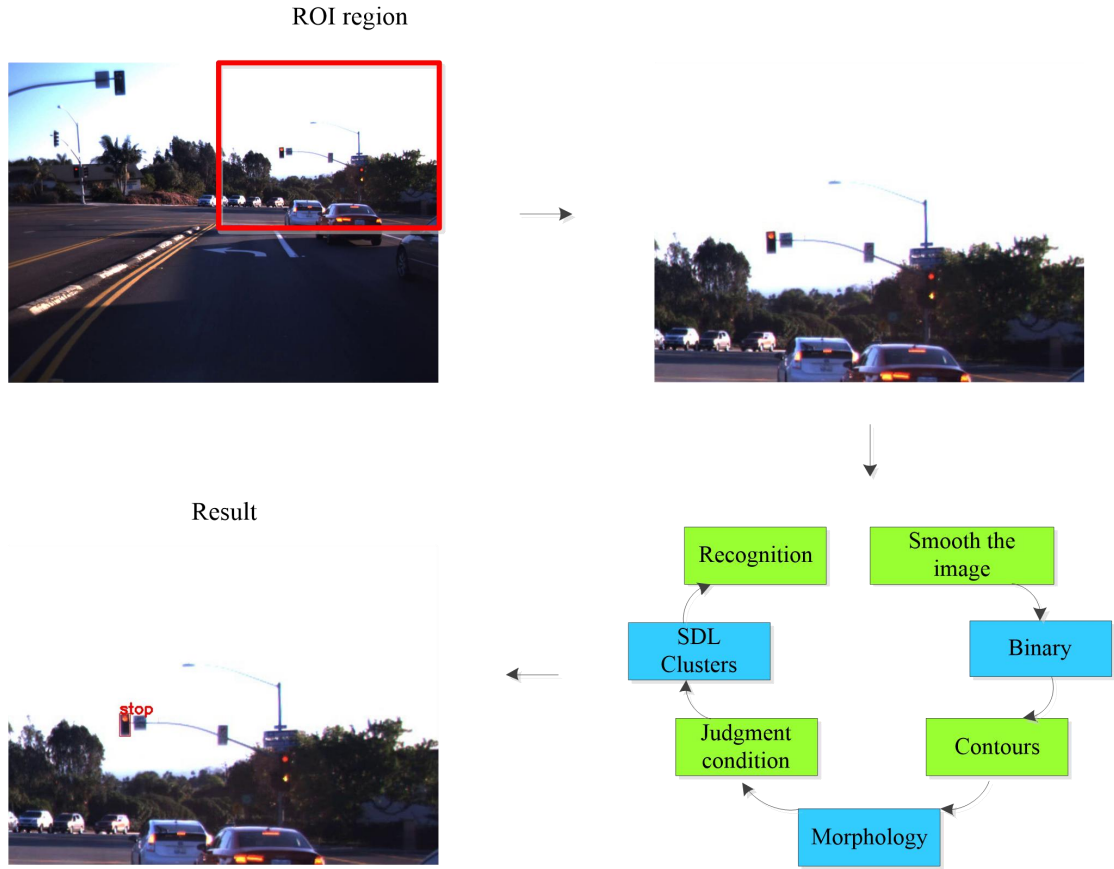

Fig. 6 flow chart of traffic light recognition

For the daytime traffic light images in LISA traffic light dataset, the pixel information corresponding to traffic lights in several images is extracted. Because of the instability of RGB color space, it is transformed into HSV space. For this part of the image, feature vectors are extracted according to different directions, and each feature vector has 18 feature values. Based on the clustering method proposed in this paper, six subspaces are finally generated. At the same time, filtering, binarization, contour, morphology and other operations are performed on the image to obtain a new image. Six subspaces are used as conditional probabilities to locate and recognize traffic lights.

Table 2 Comparison of this method and deep learning method

| Model | Data | Method | Evaluating Indicator(%) | | |
|---|---|---|---|---|---|
| | | | accuracy rate | recall rate | PR curve area |
| [26] | LISA | Fast-RCNN | 82.95 | 89.4 | — |
| [27] | LISA | Yolov3 | — | — | 90.49 |
| [28] | LISA | CNN | 96.6 | 66.6 | — |
| Our | LISA | the SDL model | 99.03 | 91 | — |

When dealing with the same LISA traffic light dataset, different models have different effects, as shown in Table 2. Although a high recall rate was obtained in reference [26], the accuracy rate was relatively low, only 82.95%. However, the reference [30] has obtained a high accuracy rate, but the recall rate has decreased significantly, which indicates that the model has serious missing inspection phenomenon, and there will be inestimable safety risks for the detection and identification of actual roads. Because the pedestrian indicator does not affect the normal operation of vehicles, it is not calculated when the correct rate and missing detection rate are counted. In this paper, through testing the daytime of LISA traffic light dataset, the accuracy rate is 99.03%, and the recall rate is 91%. Because the processing speed in CPU environment is 14fps, it can meet the real-time road detection and recognition.

## 4 Conclusion

For the LISA traffic light dataset, this paper uses the probability space clustering algorithm based on the SDL model, using less training samples, and achieves good results. Through the SDL model probability space clustering method to obtain multiple feature vectors as conditional probability information, fusion image processing related algorithms, to achieve traffic light signal location and recognition. The experimental results show that the running time of the method is fast, and the accuracy and recall rate are 99.03% and 91% respectively. In future, the probability scale self organizing method will be applied in the actual road to assist the automatic driving system.

**Funding Statement**

Our work is supported by Digital Media Art, Key Laboratory of Sichuan Province, Sichuan Conservatory of Music, Project No. :21DMAKL01; Supported by the first batch of industry-university cooperation collaborative education project funded by the Ministry of Education of the People's Republic of China, 2021, Project No. :202101071001; Supported by Minjiang College 2021 school-level scientific research project funding, Project No. :MYK21011.

**Reference**
[1]Aggarwal C C, Hinneburg A, Keim D A. On the surprising behavior of distance metrics in high dimensional spaces. International Conference on Database Theory, 2001, 1, 420-434.
[2]Beyer K, Goldstein J, Ramakrishnan R., Shaft U. When is nearest neighbor meaningful? In: International Conference on Database Theory, 1999, 217-235.
[3] J Macqueen. Some methods for classifification and analysis of multivariate observations. In 5-th Berkeley Symposium on Mathematical Statistics and Probability, 1967, 281-297, .
[4] Andrew Y Ng, Michael I Jordan, Yair Weiss. On spectral clustering: Analysis and an algorithm. In T G Dietterich, S Becker, and Z Ghahramani, editors, NeurIPS, pages 849-856. MIT Press, 2002.
[5] Tian Zhang, Raghu Ramakrishnan, Miron Livny. Birch: An effifience data clustering method for very large databases. SIGMOD Conference, 1996.
[6]Xing E P, Jordan M I, Russell S J, Ng A Y. Distance metric learning with application to clustering with side-information. Advances in neural information processing systems, 2003, 521-528.

[7]Bouzas D, Arvanitopoulos N, Tefas A. Graph embedded nonparametric mutual in formation for supervised dimensionality reduction. IEEE transactions on neural networks and learning systems, 2015, 26(5), 951-963.

[8]Nikitidis S, Tefas A, Pitas I. Maximum margin projection subspace learning for visual data analysis. IEEE Transactions on Image Processing, 2014, 23(10), 4413-4425.

[9]Passalis N, Tefas A. Dimensionality reduction using similarity-induced embeddings. IEEE transactions on neural networks and learning systems, 2017.

[10]Ding C, He X. K-means clustering via principal component analysis. Proceedings of the twenty-first international conference on Machine learning, 2004, 29.

[11]Ding C, Li T. Adaptive dimension reduction using discriminant analysis and k-means clustering. Proceedings of the 24th international conference on Machine learning, 2007, 521–528.

[12]Boutsidis C, Zouzias A, Mahoney M W, Drineas P. Randomized dimensionality reduction for k-means clustering. IEEE Transactions on Information Theory, 2015, 61(2), 1045–1062.

[13] D Pathak, P Krahenbuhl, J Donahue, T Darrell, A A Efros. Context encoders: Feature learning by inpainting. In 2016 IEEE Conference on Computer Vision and Pattern Recognition (CVPR), 2016, 2536–2544.

[14] Richard Zhang, Phillip Isola, Alexei A Efros. Colorful image colorization. European Conference on Computer Vision (ECCV), 2016, 9907: 649–666.

[15] Mehdi Noroozi, Paolo Favaro. Unsupervised learning of visual representations by solving jigsaw puzzles. In Bastian Leibe, Jiri Matas, Nicu Sebe, and Max Welling, editors, European Conference on Computer Vision (ECCV), 2016, 9910: 69–84.

[16] Mehdi Noroozi, Hamed Pirsiavash, and Paolo Favaro. Representation learning by learning to count. In ICCV, pages 5899–5907. IEEE Computer Society, 2017.

[17] Spyros Gidaris, Praveer Singh, Nikos Komodakis. Unsupervised representation learning by predicting image rotations. In ICLR. OpenReview.net, 2018.

[18] Mathilde Caron, Piotr Bojanowski, Armand Joulin, Matthijs Douze. Deep clustering for unsupervised learning of visual features. In Vittorio Ferrari, Martial Hebert, Cristian Sminchisescu, and Yair Weiss, editors, European Conference on Computer Vision (ECCV), 2018, 11218: 139–156.

[19] Morten Bornø Jensen, Mark Philip Philipsen, Andreas Møgelmose, Thomas B Moeslund, Mohan M Trivedi. Vision for Looking at Traffic Lights: Issues, Survey, and Perspectives. IEEE Transactions on Intelligent Transportation Systems, 2015.

[20] Mark Philip Philipsen, Morten Bornø Jensen, Andreas Møgelmose, Thomas B Moeslund, Mohan M Trivedi. Learning Based Traffic Light Detection: Evaluation on Challenging Dataset. 18th IEEE Intelligent Transportation Systems Conference, 2015.

[21] GU Z C, Liang Y, Zhang Z X. The Modeling of SDL Aiming at Knowledge Acquisition in Automatic Driving. arXiv: 1812.03007v1[cs.AI] 7 Dec 2018.

[22] GU Z C, Dong L. Distance Formulas Capable of Unifying Euclidian Space and Probability Space. arXiv: 1801.01972v1[cs.AI] 6 Jan 2018.

[23] George Siogkas, Evangelos Skodras, Evangelos Dermatas. Traffic Lights Detection in Adverse Conditions Using Color, Stmmetry And Spationtemporal Information. International Conference on Computer Vision Theory and Applications (VISAPP), 2012, pp. 620-627.

[24] S. Sooksatra, T. Kondo, Red traffic light detection using fast radial symmetry transform, International conference on Electrical Engineering/Electronics, Computer, Telecommunications

and Information Technology, 2014. pp. 1–6.

[25] George Siogkas, Evangelos Skodras, Evangelos Dermatas. Traffic Lights Detection in Adverse Conditions Using Color, Stmmetry And Spationtemporal Information. International Conference on Computer Vision Theory and Applications (VISAPP), 2012, pp. 620-627.

[26] V Haltakov, J Mayr, C Unger, S Ilic. Semantic segmentation based traffic light detection at day and at night, German Conference on Pattern Recognition, 2015. pp. 446–457.

[27] Y Zhang, J Xue, G Zhang, Y Zhang, N Zheng. A multi-feature fusion based traffic light recognition algorithm for intelligent vehicles. Proceedings of the 33rd Chinese Control Conference. 2014, 4924-4929.

[28] Lu Y F, Lu J M, Zhang A H, Hall P. Traffic signal detection and classification in street views using an attention model. Computational visual media, 2018,4(3):253-266.

[29] Jensen M B, Nasrollahi K, Moeslund T B. Evaluating State-of-the-Art Object Detector on Challenging Traffic Light Data. 2017 IEEE Conference on Computer Vision and Pattern Recognition Workshops (CVPRW), 2017, 9-15.

[30] Ouyang Z C, Niu J W, Liu Y, Guizani M. Deep CNN-based Real-time Traffic Light Detector for Self-driving Vehicles. IEEE Transactions on Mobile Computing. 2019, 19(2): 300-313.